\documentclass[11pt]{article}

\usepackage[utf8]{inputenc}
\usepackage[margin=1in]{geometry}
\usepackage{amsmath,amssymb}
\usepackage{graphicx}
\graphicspath{{figs/}{./}}
\usepackage{float}              
\usepackage[round]{natbib}      
\usepackage[hidelinks]{hyperref}
\usepackage{url}
\usepackage{booktabs}

\title{The Importance of Phase in Neural Representations:\\
An Internal Oppenheim--Lim Test of Image Classifiers}
\author{Alper Yıldırım}
\author{Alper Yıldırım\thanks{\href{mailto:yildirim.alper.dev@gmail.com}{yildirim.alper.dev@gmail.com}}}
\date{}

\begin{document}
\maketitle

\begin{abstract}
\citet{oppenheim1981importance} showed that natural images stay recognizable when
reconstructed from their Fourier phase alone, while the magnitude carries little of
their identity. We ask whether trained image classifiers reproduce this asymmetry
inside their hidden layers, and we test it causally: given two images, we transplant
the phase of one onto the magnitude of the other at a chosen layer and record which
image the prediction follows. In PRISM2D, GFNet, and ViT-B/16 the prediction follows
the phase or sign donor, and deleting all image-specific magnitude barely moves
accuracy, so identity rides on phase while image-specific magnitude is largely
dispensable to the readout. ResNet-50
at first seems to break the pattern, because transplanting sign after its ReLUs does
nothing; a fair intervention before the ReLU reveals a strong latent sign code in the
late blocks, and a DC-only control shows the readout consumes a channel-wise spatial
average. Controls rule out the trivial case in which magnitude simply stops depending
on the image. The architectures therefore share a phase/sign identity code but expose
it in different bases, set by rectification and readout geometry, which gives a
mechanistic account of the texture--shape gap between CNNs and attention models.
\end{abstract}


\section{Introduction}

A Fourier transform splits a signal into magnitude and phase. For natural images
these two components are far from symmetric in what they carry. In a now-classic
demonstration, \citet{oppenheim1981importance} swapped the magnitude and phase
spectra of two images and observed that the reconstruction is recognizable as the
\emph{phase} donor: phase encodes the edges, contours, and spatial arrangement that
make an image intelligible, while magnitude encodes a comparatively
content-agnostic energy profile. The asymmetry is fundamental enough that a signal
can, under broad conditions, be reconstructed from phase alone up to a scale factor.

Modern image classifiers are trained end-to-end on pixels with no Fourier
transform imposed on them, yet the question of whether they recover this
phase-centrism has only been studied at the \emph{input}. \citet{chen2021apr}
recombine the amplitude of one image with the phase of another as a data
augmentation and find that CNNs over-rely on amplitude relative to humans; a
parallel line of work in domain adaptation treats the amplitude spectrum as
``style'' and the phase spectrum as ``content,'' swapping amplitudes to synthesize
cross-domain data \citep{yang2020fda}. These methods establish that the
phase--magnitude asymmetry is visible to networks, but they operate strictly in
pixel space and as augmentation; they do not ask where, \emph{inside} a trained
model, the decision-relevant information lives, nor whether it can be moved. In
language models, by contrast, frequency-domain structure has already been found
\emph{inside} the representations: pre-trained transformers compute addition using
Fourier features in their hidden states, and represent numbers on a periodic helix
they manipulate to add, both established through causal interventions
\citep{zhou2024pretrained,kantamneni2025trigonometry}. We ask the analogous
question for vision.

\paragraph{This work.}
We lift the Oppenheim--Lim experiment from pixels into the latent stream and run it
as a causal intervention rather than an augmentation, in the spirit of activation
patching and causal mediation analysis
\citep{vig2020causal,meng2022rome}. Given a pair of
different-class images $A$ and $B$, we construct a chimeric hidden state that keeps
the \emph{magnitude} of $A$'s intermediate features but takes the \emph{phase-like}
component from $B$, inject it at a chosen layer, finish the forward pass, and record
whether the model predicts the magnitude donor or the phase donor. The phase-like
component is exact where the stream is complex (the per-channel complex phase),
becomes the spatial-spectral phase when we Fourier-transform a real feature map over
its spatial grid, and degenerates to the per-unit \emph{sign} in a signed real
stream---the only ``phase'' a real activation has. Sign therefore serves as the
common probe across all architectures, with complex and spectral phase as the richer
special cases.

We study four architectures spanning the relevant design axes. PRISM2D adapts the
PRISM architecture of \citet{yildirim2026prism}, a 1D complex-valued model with
phase-preserving activations and a unit-norm phase constraint, to 2D images. It
maintains a genuinely complex latent stream and gives the cleanest testbed for a
semantic role of phase. Unlike complex-valued networks built for inherently complex
physical signals such as MRI or SAR \citep{trabelsi2018deepcomplex,vasudeva2022covegan,viger2025sar},
PRISM2D operates on ordinary optical images. GFNet \citep{rao2021gfnet} is a
real-valued spectral model that filters in the Fourier domain but carries a
signed real residual stream. ViT-B/16 \citep{dosovitskiy2021vit} and ResNet-50
\citep{he2016resnet} are standard, non-spectral, off-the-shelf classifiers included
to test generality.

\paragraph{Findings.}
First, the phenomenon is real and not architecture-bound: in all four models the
prediction can be driven to follow the phase/sign donor, and substituting every
image's magnitude with a batch-mean magnitude---deleting all image-specific
magnitude information---leaves accuracy close to baseline for the spectral models and
for ViT. Identity rides on the phase-like component, and the readout is largely
magnitude-invariant. Second, the effect is not the trivial consequence of magnitude
becoming uninformative with depth: magnitudes remain measurably image-specific, the
chimera stays far from the phase donor in feature space, and destroying phase
outright (random phase, or unit magnitude) collapses accuracy to chance. A DC-only
control further shows that the transplanted identity lives in genuine non-DC spatial
structure in early and middle layers, not in the global mean (brightness) of the
feature map.

Third, and most informative, the \emph{route} to this phase code is governed by
architecture and readout geometry, not shared across models. ViT, whose class token
can carry a distributed code from the outset, commits to a sign-based identity code
in its very first block and strengthens it monotonically with depth. The spectral
models consolidate identity into phase by early-to-middle depth. ResNet is the
striking case: a naive sign transplant on its rectified stream appears to do nothing,
which would suggest a purely magnitude-based code---but this is an artifact of
rectification, which displaces sign information into magnitude. A \emph{fair}
intervention applied before the final ReLU of each residual block reveals that late
ResNet blocks follow the transplanted sign almost as strongly as ViT. Reading the
three probes together---decaying spatial-spectral phase, rising pre-activation sign,
and a late handoff of identity into the spatial-mean (DC) channel that global average
pooling consumes---shows that ResNet \emph{transports} class identity from
distributed spatial structure in early layers into a pooling-readable channel code in
its final blocks.

We therefore propose a single account: image classifiers of otherwise very different
design converge on a phase/sign-dominant identity code by the time of readout, and
differ chiefly in \emph{when} (onset depth) and in \emph{which basis} that code is
expressed---an ordering plausibly shaped by readout geometry (a class token versus
global pooling) and by layer inductive bias (spectral versus convolutional). This
reframes the long-observed behavioral gap between texture-biased CNNs and more
shape-/human-aligned attention models
\citep{geirhos2019texture,tuli2021human} as a difference in where and how readily
identity is committed to phase, rather than a categorical phase-versus-magnitude
split.

\paragraph{Contributions.}
\begin{itemize}
  \item We recast the Oppenheim--Lim phase-importance phenomenon as a \emph{causal,
  layer-resolved intervention} on latent representations, with the per-unit sign as a
  probe that applies uniformly to complex, spectral, and standard real-valued
  networks.
  \item We show across four architectures that class identity is carried by the
  phase-like component of intermediate features and that the readout is largely
  magnitude-invariant, supported by controls (random-phase, unit-magnitude,
  mean-magnitude, and DC-only) that exclude trivial explanations.
  \item We demonstrate that the sign code of rectified ConvNets is \emph{latent}:
  it is hidden by ReLU and recoverable only by a pre-activation intervention, and we
  use this to map ResNet's late-depth handoff of identity into the pooling-readable
  DC/channel code.
  \item We give a unified, readout-aware account in which all four architectures
  reach a phase/sign-dominant identity code by readout but along architecture-specific
  routes, and connect it to the texture--shape literature.
\end{itemize}

\section{Method}
\label{sec:method}
We evaluate four trained classifiers on the same protocol. ImageNet-100 has 100
classes, with 126{,}689 training and 5{,}000 validation images. PRISM2D (depth 10,
width 256) and GFNet-Ti (depth 12, width 256) are trained from scratch on this set.
They are matched in capacity, about 7M parameters each, and in clean accuracy, about
78\% top-1, so comparisons between them isolate the complex against the real spectral
design. ResNet-50 and ViT-B/16 are larger off-the-shelf models, about 25M and 86M
parameters, that use the public ImageNet-1k weights; we restrict their logits to the
100 ImageNet-100 classes through a name map, which keeps chance at 1\% and their
clean accuracy near 93\%. They serve as generality checks, not capacity-matched
competitors. Together the four span four architecture families, complex, real
spectral, convolutional, and attention, and roughly an order of magnitude in scale,
from about 7M to 86M parameters, so a shared finding is unlikely to be an artifact of
one architecture or one model size. All interventions are applied at inference only. The forward pass runs
in \texttt{bfloat16} autocast, while every Fourier transform and phase operation is
computed in single precision.

\subsection{Neural Oppenheim--Lim intervention}
\label{sec:method-intervention}

We first sample image pairs of \emph{different} classes. We shuffle the validation
set with a fixed seed and walk it in order, keeping each consecutive pair
$(x_A, x_B)$ only when $y_A \neq y_B$, until we have 2488 different-class pairs, that
is 4976 evaluation images.
Pairs are laid out interleaved in the batch, so one forward pass scores both members.

The original Oppenheim--Lim experiment takes two images, swaps their Fourier
magnitude and phase, and inverts the transform. The reconstruction is read as the
phase donor, not the magnitude donor \citep{oppenheim1981importance}. We move this
idea from the pixels to a hidden layer.

Fix a layer and call its output $h^{(\ell)}(x)$, the representation of image $x$ at
site $\ell$. This is usually a grid of numbers, one per spatial location and feature
channel, not a single vector. We call one scalar entry a \emph{coordinate}. In
PRISM2D a coordinate is one location and one channel, and it is complex; in the real
models it is an ordinary real number.

Every coordinate has two parts: a \emph{magnitude} $|h|$, its positive size, and a
\emph{phase} $\phi(h)$, the unit-size part that says which way it points. For a real
number the phase is just its sign. For a complex number it is an angle. The neural
Oppenheim--Lim chimera keeps one image's magnitudes and pastes in the other's phases,
\[
  \tilde h_A = |h_A^{(\ell)}|\,\phi(h_B^{(\ell)}),
  \qquad
  \tilde h_B = |h_B^{(\ell)}|\,\phi(h_A^{(\ell)}).
\]
so $\tilde h_A$ has $A$'s strengths but $B$'s structure. We put $\tilde h$ back into
the layer, finish the forward pass, and read the predicted class. Each pair gives two
chimeras at once. Setting $\ell$ to the input pixels and $\phi$ to the Fourier phase
recovers the 1981 experiment, which we report as a pixel-space baseline. The point of
this paper is to place $\ell$ inside the network.

\subsection{Phase-like components in different representation spaces}
\label{sec:method-phase}

The phase-like part $\phi$ takes three forms, depending on the stream.

\paragraph{Complex stream (channel phase).} When the state is complex,
$h \in \mathbb{C}^{H_\ell \times W_\ell \times D}$, the phase is exact and
coordinate-wise, $\phi(h) = h / |h|$. The chimera keeps $A$'s coordinate-wise modulus
and takes $B$'s coordinate-wise angle. This is the richest case and applies to PRISM2D.

\paragraph{Real feature map (spatial-spectral phase).} For a real feature map we use
the analyst's Fourier transform over the spatial grid. Let
$H = \mathcal{F}_{\mathrm{2D}}(h)$. We replace the spectrum with
$|H_A|\,H_B / |H_B|$ and invert, $\tilde h_A = \mathcal{F}_{\mathrm{2D}}^{-1}(|H_A|\,H_B/|H_B|)$.
This mirrors the pixel-space experiment one layer up. The transform is ours as the
analyst, not a computation the model performs, exactly as in the 1981 setup.

\paragraph{Real vector (sign).} A real scalar activation has only two phases, $0$ and
$\pi$, so its phase-like part is its sign, $\phi(h) = \operatorname{sign}(h)$. The
chimera is $\tilde h_A = |h_A|\,\operatorname{sign}(h_B)$. Sign is defined for every
real stream, so it is the common probe across all four models; complex and
spatial-spectral phase are the richer special cases where the architecture allows
them.

\subsection{Scoring: donor-following metrics}
\label{sec:method-scoring}

A chimera at $A$ carries $A$'s magnitude and $B$'s phase. We call $A$ the magnitude
donor and $B$ the phase donor. With $\hat y_A = \arg\max f(\tilde h_A)$ we report
three rates over all chimeras:
\[
  \text{follow-phase} = \Pr[\hat y_A = y_B], \quad
  \text{follow-magnitude} = \Pr[\hat y_A = y_A], \quad
  \text{other} = 1 - \text{(the two)}.
\]
For the degenerate real case we read follow-phase as follow-sign. Chance is 1\%.
The clean model with no intervention gives follow-magnitude equal to its accuracy.
We sweep $\ell$ over every layer or block and plot these rates against depth.

\subsection{Controls against trivial explanations}
\label{sec:method-controls}

Follow-phase could be trivial if late-layer magnitudes stopped depending on the
image, since then $|h_A|\,\phi(h_B)$ would already be close to $h_B$. We test this
from three directions.

\paragraph{Phase necessity.} We destroy phase while keeping magnitude (random phase),
and we destroy magnitude while keeping phase (unit magnitude). If random phase
collapses accuracy toward chance, phase carries the class signal. These apply the
operation at all sites and report self-label accuracy.

\paragraph{Magnitude sufficiency.} We replace each image's magnitude with the
batch-mean magnitude pattern and keep its own phase or sign. This deletes all
image-specific magnitude information and imports none from a partner. If self-label
accuracy stays near baseline, the readout does not rely on image-specific magnitude.

\paragraph{Geometry.} Per layer we measure the across-image magnitude correlation
$\operatorname{corr}(|h_A|,|h_B|)$, and the cosine between the chimera and the phase
donor, with the cosine between the two clean states as a floor. The trivial hypothesis predicts the magnitude correlation approaches 1 and the chimera
approaches the phase donor. Observed values stay well below 1, so we reject it.

\paragraph{Spatial locus (ResNet).} For the spatial-spectral case we swap only the
DC coefficient, and separately everything except DC. This separates a global mean,
or brightness, effect from genuine spatial structure, and it identifies what the
global-average-pooling readout consumes.

\subsection{Model-specific intervention sites}
\label{sec:method-sites}

We intervene on the residual stream, that is, on the state entering a layer or block,
unless noted otherwise.

\paragraph{PRISM2D.} The residual stream is complex. We transplant the per-channel
complex phase at the input to layer $\ell$ (the channel-phase site). The model also
computes an internal 2D Fourier transform inside each gated harmonic convolution; we
can swap phase there, before the learned spectral filter, at a single layer or
cumulatively over layers $0..\ell$ (the branch site). Readout is a phase
normalization, a complex-to-real bridge, a spatial mean, and a linear head.

\paragraph{GFNet-Ti.} The residual stream is real. Each global filter computes an
\texttt{rfft2} over the token grid; we swap phase there before the learned filter
(the branch site). On the stream we apply the analyst spatial-spectral swap by
reshaping tokens to the grid, transforming, swapping, and inverting (the
stream-spectral site), and the per-token sign swap (the sign site). Readout is a
layer normalization, a token mean, and a head \citep{rao2021gfnet}.

\paragraph{ResNet-50.} States are real feature maps. Between blocks we apply the
sign swap and the analyst spatial-spectral swap over the spatial axes. Because
post-activation maps are non-negative, the between-block sign is nearly degenerate.
We therefore also intervene inside each bottleneck on the signed pre-activation, that
is, on the residual sum before the block's final ReLU (the pre-ReLU sign site), which
is the fair sign test. We add the DC-only and non-DC swaps, including the site
immediately before global average pooling. Readout is the pooling, the linear layer,
and the restricted logits.

\paragraph{ViT-B/16.} States are real token sequences. The sign swap is applied to
all tokens including the class token. The spatial-spectral swap is applied to the
patch tokens only, reshaped to a $14\times 14$ grid, leaving the class token
untouched. Readout uses the class token after the final layer normalization. Because
the readout reads only that token, a late patch-token swap can leave the prediction
unchanged through readout locality rather than phase robustness, and we flag the
affected layers.

\section{Results}

\subsection{Phase decides the class in every architecture}
\begin{figure}[H]
  \centering
  \includegraphics[width=\linewidth]{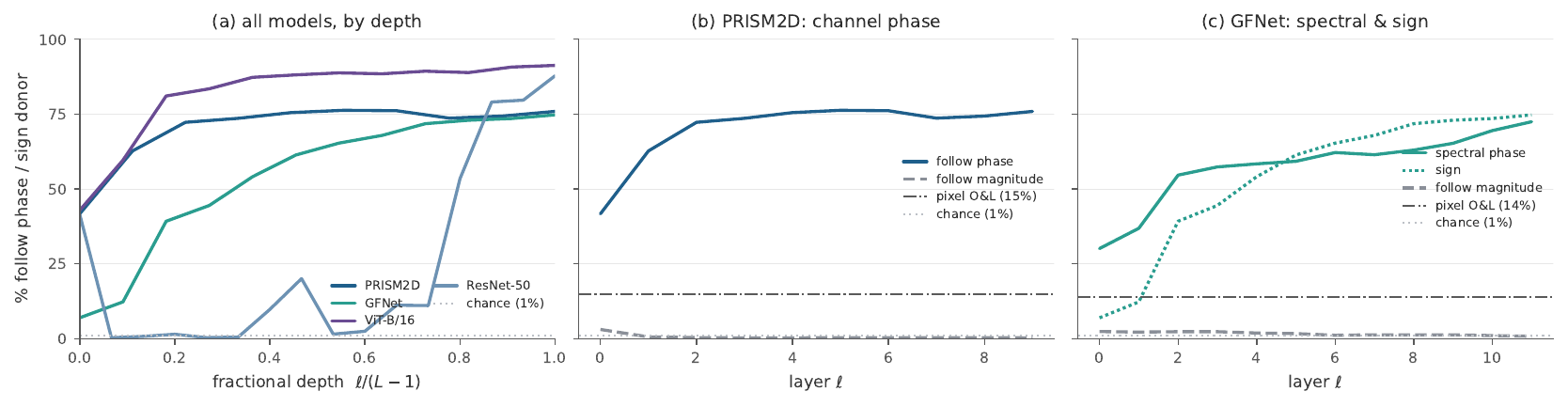}
  \caption{Phase and sign carry class identity. (a) Fraction of chimeras whose
  prediction follows the phase or sign donor, against fractional depth; all four
  models climb to a high plateau but commit at different depths. (b) PRISM2D channel
  phase and (c) GFNet spatial-spectral phase and sign drive the prediction to the
  phase donor while magnitude-following stays at the floor. The same swap on raw
  pixels (dash-dot) barely moves the prediction, so the code is internal.}
  \label{fig:overview}
\end{figure}

We sweep the intervention layer and record how often the prediction follows the
phase or sign donor. Figure~\ref{fig:overview}(a) shows the result. In all four models the
curve rises from chance to a high plateau. By the last block ViT reaches 91\%, ResNet
88\%, PRISM2D 76\%, and GFNet 75\%. The models differ in when the code appears. ViT
commits in its first block. PRISM2D and GFNet build up over early and middle depth.
ResNet stays near the floor until its last few blocks and then jumps. The shared
endpoint is the main finding: by the time of readout, the phase or sign of the
features decides the class.

PRISM2D and GFNet are our matched testbeds, so we look at them closely
(Figure~\ref{fig:overview}b,c). Transplanting channel phase in PRISM2D drives
follow-phase to about 76\% by layer~2 and holds it there, while follow-magnitude
falls below 1\%. GFNet behaves the same way through two probes: its spatial-spectral
phase reaches 72\% and its per-unit sign reaches 75\%, again with magnitude-following
at the floor. The dash-dot line is the control that matters most. The same swap on
raw pixels moves the prediction only about 14\%. Inside the network it moves it five
times as much. The phase code is something the models build, not something inherited
from the pixels.

\subsection{Controls}
\label{sec:controls}
\begin{figure}[H]
  \centering
  \includegraphics[width=\linewidth]{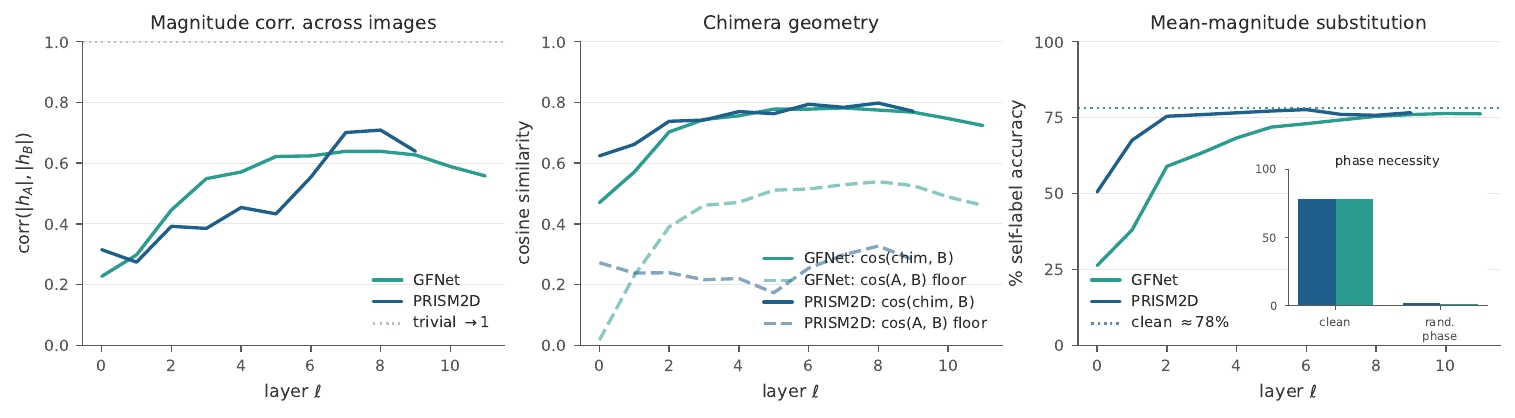}
  \caption{Controls. From left: magnitudes stay image-specific (correlation well
  below 1); the chimera is far from the phase donor in feature space; and deleting
  image-specific magnitude leaves accuracy near baseline. The inset in the third panel
  shows the opposite ablation, where destroying phase collapses accuracy to near
  chance. Image-specific magnitude is largely dispensable, while destroying phase
  collapses accuracy.}
  \label{fig:controls}
\end{figure}

A skeptic could argue that late magnitudes stop depending on the image, so the
chimera is already the phase donor and following it is trivial. Four checks rule this
out (Figure~\ref{fig:controls}).

\paragraph{(1) Magnitudes stay image-specific.} The across-image magnitude
correlation rises with depth but stays near 0.6, never close to 1. If magnitudes were
shared across images, it would approach 1.

\paragraph{(2) The chimera is not the donor.}
The phase/sign transplant does not simply reconstruct the donor representation.
For all three probes, the chimera--donor cosine is tied algebraically to the
magnitude overlap, \(\cos(|h_A|,|h_B|)\), and is therefore structurally inflated
by magnitude non-negativity and representation anisotropy
(App.~\ref{app:cosine}). Read against the same-layer \(\cos(A,B)\) floor, the
chimera remains geometrically distinct from the donor. We therefore place the
evidential weight on the prediction flip rather than on the absolute cosine:
the readout calls the chimera the donor's class even though the chimera is not
the donor state.

A donor-replacement control supports this prediction-based reading: holding the
magnitude donor \(A\) fixed and replacing the phase/sign donor \(B\) by a third
image \(C\) redirects the prediction from \(B\) to \(C\)
(App.~\ref{app:donor-replacement}).

\paragraph{(3) Image-specific magnitude is dispensable.} Replacing each image's
magnitude with the batch-mean magnitude deletes all image-specific magnitude, yet
accuracy stays near the clean 78\%.

\paragraph{(4) Phase is necessary.} Destroying phase instead, by randomizing it,
collapses accuracy to about 2\%, near chance.

\noindent Together: phase is necessary, while the image-specific magnitude pattern is
largely dispensable to the readout.

\subsection{ResNet: ReLU hides the sign, and pooling reads the DC term}
\begin{figure}[H]
  \centering
  \includegraphics[width=0.72\linewidth]{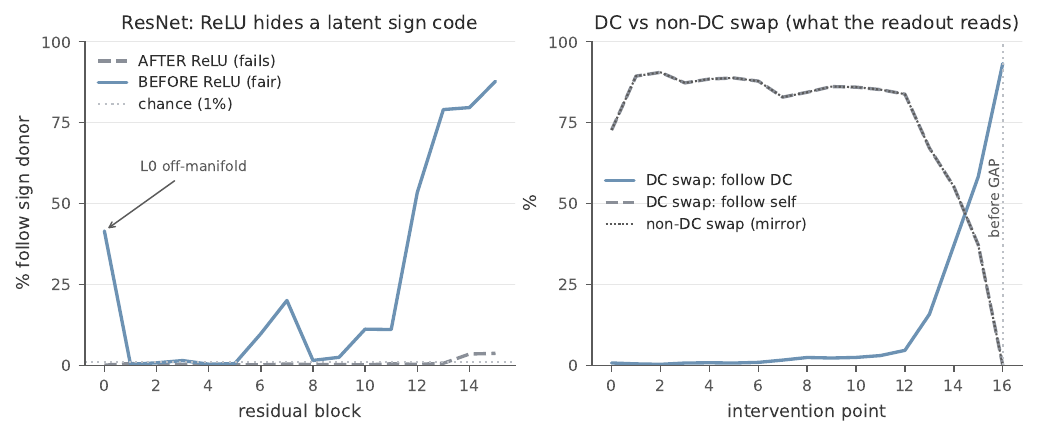}
  \caption{ResNet mechanism. Left: sign swapped after the ReLU does nothing; swapped
  before the ReLU it transplants the class in late blocks. Right: only the DC bin
  matters just before global average pooling, where swapping it transplants the class
  completely. The block-0 point is off-manifold (48\% other).}
  \label{fig:resnet}
\end{figure}

ResNet first looks like the exception (Figure~\ref{fig:resnet}, left). Swapping the
sign of its activations does almost nothing; the grey curve stays at the floor. The
reason is the ReLU. After it, every value is non-negative, so the sign is always
positive and there is nothing to swap. Move the swap to just before the ReLU, where
values are still signed, and the late blocks follow the sign up to 88\%. The sign code
was there all along, hidden by rectification. The right panel explains the readout.
Global average pooling is the spatial mean, which is exactly the DC Fourier term.
Swapping only the DC bin does nothing until the last few blocks, then transplants the
class completely right before pooling, at 93\%, while swapping everything except DC is
the mirror. So identity starts in spatial structure and is folded into the DC channel
that pooling reads.

\subsection{Generality: ViT is phase-coded, and ResNet's full picture}
\begin{figure}[H]
  \centering
  \includegraphics[width=0.72\linewidth]{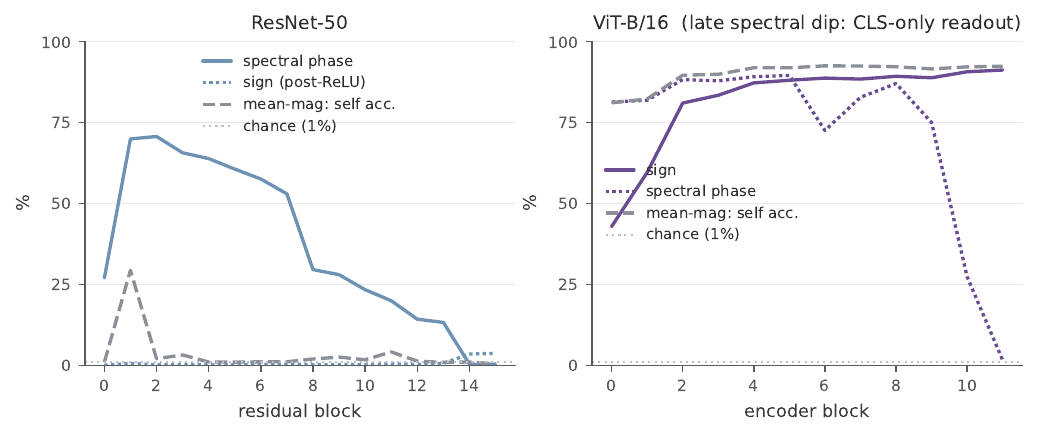}
  \caption{Generality. ViT (right) follows the sign throughout and ignores magnitude.
  ResNet (left) carries the class in spatial-spectral phase early, loses it in
  post-ReLU sign, and depends on magnitude after rectification. The late ViT spectral
  dip is readout locality from the class-token readout, not phase robustness.}
  \label{fig:generality}
\end{figure}

The standard models confirm the pattern and sharpen the contrast
(Figure~\ref{fig:generality}). ViT follows the sign from the first block up to 91\%,
and deleting image-specific magnitude barely touches accuracy, which stays above
90\%. ViT is phase and sign coded throughout, and magnitude invariant. ResNet tells a
depth story. Its spatial-spectral phase carries the class strongly early, about 70\%
near block~2, then decays as identity moves into the poolable channel code. Its
post-ReLU sign carries nothing, the same artifact as before. And mean-magnitude
deletion is catastrophic, near chance, because after rectification the magnitude is
where the signal sits. The one caveat is the late dip in ViT's spectral curve. That
swap touches only patch tokens, but the readout uses the class token, which has
already gathered the class by late layers, so the dip is readout locality, not phase
robustness.

\section{Limitations}
We study vision models. The transplant itself is modality-general, since it needs only
a magnitude and a phase part, which any complex or spectral representation provides.
Audio is the natural next case, because a spectrogram carries magnitude and phase
directly; the sign form of the probe could in principle reach further, though the clean
magnitude--phase split is most natural for signals with a spectral representation. We
keep the scope to vision on purpose. Images give the cleanest mechanistic test, with
well understood spectral structure and controlled labels, and isolating them keeps the
claims sharp and marks a focused next step.

Phase is not a single object here. It is a complex channel phase in PRISM2D, a
spatial-spectral Fourier phase in GFNet and the ResNet probe, and a real sign in ViT
and the pre-ReLU ResNet test. These are related but not identical. The convergence we
report is across these operationalizations, not within one definition, so phase should
be read as a family of related codes rather than a single quantity.

Our models are not matched on training data. PRISM2D and GFNet are trained on
ImageNet-100, while the ViT and ResNet are ImageNet-1k models read out on the same 100
classes, so the cross-model comparison mixes a difference in training scale with the
architectural differences we study. The within-model depth trends, which carry most of
our claims, are not affected by this. One further caveat: the ViT spectral probe
perturbs only patch tokens, so its late dip reflects the class-token readout rather
than a loss of phase coding.

\section{Discussion}
A linear classifier reads a direction in feature space, not a length, so an identity
code that lives in the angle rather than the magnitude is close to what the readout
already rewards. This may be why the four architectures converge on a phase or sign
code and differ mainly in how early they reach it. Angular and periodic codes turn up
elsewhere in trained networks, such as Fourier features and helical number
representations \citep{zhou2024pretrained, kantamneni2025trigonometry}, so an angular
identity code is not surprising.

Read geometrically, magnitude is a radius and phase is an angle, and our controls say
identity is the angular coordinate. Substituting a generic magnitude keeps the
representation on the data manifold and preserves accuracy, while randomizing phase
leaves it and collapses accuracy to chance. Whether the concept manifolds recovered by
sparse autoencoders are organized along these angular directions is an open question
\citep{bhalla2026sparse}.

The split also lines up with the texture-versus-shape gap, where magnitude tracks
global energy and texture and phase tracks spatial configuration and shape, giving a
frequency-domain handle on that behavioral literature \citep{geirhos2019texture,
tuli2021human}. This mapping is clean for the spatial-spectral phase and the DC term,
and is only a loose analogy for the channel-sign code.

\paragraph{Future work.} Rectified networks hide the code and recover it only in their
last blocks, while complex and spectral models express it early. Architectures that
represent phase explicitly may therefore carry a useful inductive bias for these tasks,
and building phase-based models, or parameterizing the residual stream around an
explicit phase code, is a natural next step. Testing whether such a bias improves
sample efficiency or robustness, and whether it aligns with the manifold structure
above, would connect the two threads.

\clearpage
\section*{Acknowledgements}

The author acknowledges the use of large language models as assistive tools in the preparation of this manuscript. These tools were used for coding assistance, figure generation, debugging, and language polishing. The conception of the study, experimental design, analysis, interpretation, and final manuscript decisions were made by the author. The author takes full responsibility for the content of the paper, including any errors or omissions.

\section*{Code Availability}
Code to reproduce all experiments, including the neural Oppenheim--Lim
interventions, the control analyses, and the from-scratch PRISM2D and
GFNet-Ti training recipes, is available at
\url{https://github.com/AlperYildirim1/Oppenheim-Lim-Neural-Networks}.

\bibliography{references}

\appendix

\section{Why the chimera--donor cosine is high, and why it is benign}
\label{app:cosine}

In Section~\ref{sec:controls} we reported that the chimera sits at cosine
$0.7$--$0.8$ from the phase donor. A skeptic could read this as the chimera
being ``essentially'' the donor, so that the readout following the donor is
trivial. We argue the opposite: a cosine in this range is the \emph{expected}
geometry, for two independent reasons---one algebraic, one about the geometry of
trained representations---and the quantity that actually carries our claim is
not the cosine but the prediction-following rate.

\paragraph{An exact identity.}
For all three probes the chimera--donor cosine is not a free measurement: it is
\emph{exactly} the cosine between the two magnitude vectors. Write the real
Euclidean inner product as $\langle u,v\rangle$ (the real part, for a complex
stream), and let the chimera be $\tilde h_A = |h_A|\,\phi(h_B)$ coordinatewise.

For the real \emph{sign} probe, $\phi(h_B)=\mathrm{sign}(h_B)$ and
$\mathrm{sign}(h_{B,i})\,h_{B,i}=|h_{B,i}|$, so
\[
\langle \tilde h_A, h_B\rangle
   = \sum_i |h_{A,i}|\,\mathrm{sign}(h_{B,i})\,h_{B,i}
   = \sum_i |h_{A,i}|\,|h_{B,i}|
   = \langle |h_A|,|h_B|\rangle ,
\qquad
\|\tilde h_A\| = \|h_A\| .
\]
For the \emph{complex} probe, $\phi(h_B)=h_B/|h_B|$ and
$(h_{B,i}/|h_{B,i}|)\,\overline{h_{B,i}}=|h_{B,i}|$, giving the same sum. For the
\emph{spatial-spectral} probe the swap is applied to $H=\mathcal{F}_{2D}(h)$, and
by Parseval the inner product is preserved up to the unitary normalisation, so
the identity holds for the magnitude \emph{spectra} $|H_A|,|H_B|$. In every case
\[
\cos\!\big(\tilde h_A,\,h_B\big)\;=\;\cos\!\big(|h_A|,\,|h_B|\big).
\]
Two consequences follow. First, the chimera--donor cosine and the across-image
magnitude correlation we report are the \emph{same} quantity up to mean-centering,
which is why we report them together rather than as independent controls.
Second, because magnitudes (and magnitude spectra) are non-negative, the two
vectors lie in the positive orthant, where the cosine is bounded away from zero
and is large in high dimension. The value is therefore structurally elevated
\emph{before} any appeal to learned structure.

\paragraph{Anisotropy of trained representations.}
The hidden states of unrelated inputs are not spread evenly over direction; they
concentrate in a narrow cone, so the cosine between the representations of two
unrelated inputs is high as a matter of ambient geometry rather than shared
content \citep{ethayarajh2019contextual,gao2018representation}. This is the
representation-degeneration / anisotropy phenomenon, and \citet{godey2024anisotropy}
show it is inherent to self-attention and present in transformers trained on
modalities beyond text, which is what licenses invoking it for ViT-B/16 rather
than for language models alone. The cosine is moreover dominated by a few shared
high-variance dimensions that need not be class-discriminative, so absolute
cosine systematically overstates relatedness and is a poor measure of
representational similarity \citep{Timkey2021bark}. We use the
naming/phenomenon results only; we do \emph{not} import their language-specific
cause (rare-token gradients and tied output embeddings), which has no analogue in
our image classifiers.

\paragraph{The correct reading: gap above the in-model floor.}
Anisotropy implies that an absolute cosine cannot be read as identity-sharing;
the meaningful quantity is the gap above the baseline cosine between unrelated
states \emph{in the same model and layer}. We do not borrow this baseline from
the literature: the $\cos(A,B)$ floor plotted in Figure~\ref{fig:controls} is
exactly that baseline, measured on our own unrelated states. Read against it, the
chimera's $0.7$--$0.8$ is a gap above the floor, not proximity to the donor; the
chimera is the donor's class while remaining geometrically distinct from the
donor.

\paragraph{Why we foreground prediction-following.}
Because absolute cosine is structurally inflated (by the identity above) and
ambiently inflated (by anisotropy), we treat the geometry only as a consistency
check and place no evidential weight on the cosine itself. The load-bearing
measurement is the readout's predicted class: a sign/phase transplant changes
\emph{which class the model outputs}, an effect that lives in the decision of the
head and is outside the cone argument entirely. This also answers the natural
rejoinder that the transplant merely moves the state ``within the noise cone'':
movement inside the cone does not, on its own, change the predicted label, yet
the prediction flips to the donor. A still cleaner null---resampling the
magnitude to hold $\cos(|h_A|,|h_B|)$ fixed while breaking the donor's identity,
and checking that follow-phase still tracks the true donor---would isolate this
further; we leave a full version to future work, as the identity and the
floor comparison above already exclude the trivial explanation.

\subsection{Donor-replacement control}
\label{app:donor-replacement}
To test whether the readout follows the transplanted donor rather than the original image pair, we add a third-image control. For each different-class pair \(A,B\), we keep the magnitude donor \(A\) fixed and replace the phase/sign donor \(B\) with a third image \(C\), with \(y_C \neq y_A,y_B\):
\[
\tilde h_{AB}=|h_A|\phi(h_B),
\qquad
\tilde h_{AC}=|h_A|\phi(h_C).
\]
If identity is carried by the transplanted phase/sign pattern, then \(\tilde h_{AB}\) should be classified as \(B\), while \(\tilde h_{AC}\) should be classified as \(C\).

This is what we observe. In PRISM2D, \(\tilde h_{AB}\) follows \(B\) at \(72.2\%\) and \(75.9\%\) at layers 2 and 9, while \(\tilde h_{AC}\) follows the old donor \(B\) only \(1.4\%\) and \(2.7\%\), and follows the new donor \(C\) at \(76.4\%\) and \(84.3\%\). In GFNet at layer 11, \(\tilde h_{AB}\) follows \(B\) at \(74.7\%\), while \(\tilde h_{AC}\) follows old \(B\) at \(12.0\%\) and new \(C\) at \(74.5\%\). Thus, holding magnitude fixed, changing only the phase/sign donor redirects the predicted identity to the new donor.

\section{PRISM2D Architecture}
\label{app:arch}

PRISM2D adapts the 1D PRISM encoder of \citet{yildirim2026prism} to images.
It carries a complex residual stream $X \in \mathbb{C}^{H\times W\times D}$ over an
$H\times W$ patch grid with $D$ channels, replaces attention with \emph{Gated
Harmonic Convolutions} that mix globally through a 2D FFT, and is
phase-preserving from input to readout except for a single complex-to-real
bridge. There is no attention and no real (``particle'') stream: unlike the
hybrid models of \citet{yildirim2026prism}, the classifier uses the complex
(wave) stream only. We write a complex coordinate as $z = r e^{i\theta}$, with
magnitude $r=|z|$ and phase $\theta=\angle z$.

\paragraph{Complex tokenisation and 2D rotary position.}
A patch is embedded by a strided convolution into real features
$p \in \mathbb{R}^{H\times W\times D}$, lifted to a complex content vector
$z_t \in \mathbb{C}^{H\times W\times D}$ by a learned linear adapter, and rotated
by two unit-modulus phasors:
\begin{equation}
E = z_t \,\odot\, R_{\mathrm{pos}} \,\odot\, R_{\mathrm{dyn}}, \qquad |R_{\mathrm{pos}}|=|R_{\mathrm{dyn}}|=1 .
\end{equation}
Because each channel is already a phasor, position enters as the complex-valued
analogue of RoPE: we split the channels in half and rotate one half by the row
coordinate and the other by the column coordinate, with geometric frequencies
$\omega_k = 10000^{-k/(D/2)}$, so the per-channel angle is
$\theta^{\mathrm{pos}}_k = \mathrm{(row\ or\ col)}\cdot \omega_k$ and
$R_{\mathrm{pos}} = e^{i\theta^{\mathrm{pos}}}$. The content-dependent steering
phasor $R_{\mathrm{dyn}} = e^{i\phi_{\mathrm{steer}}}$ is predicted per token from
$p$ and normalised to unit modulus.

\paragraph{Gated Harmonic Convolution.}
Each block normalises the stream, mixes it globally in the Fourier domain with a
learned complex filter $H$, applies a complex gate, a phase-preserving
non-linearity, and a complex linear projection, with a residual connection:
\begin{align}
\tilde X &= \mathrm{PhaseNorm}(X), &
Y &= \mathcal{F}_{2D}^{-1}\!\big(\mathcal{F}_{2D}(\tilde X)\odot H\big), \\
Z &= \Phi_v(Y)\,\odot\,\Phi_g(Y), &
X_{\mathrm{out}} &= X + \mathrm{Drop}\big(W_{\mathrm{out}}\,\mathrm{ModReLU}(Z)\big).
\end{align}
Here $\mathcal{F}_{2D}$ is the 2D FFT over the spatial axes only (channels
untouched), and $\Phi_v,\Phi_g$ each map $[\mathrm{Re}\,Y \,\|\, \mathrm{Im}\,Y]$
through a linear layer whose output is reinterpreted as a complex vector. The
gate is a \emph{complex} product, so magnitudes multiply and phases add,
$|Z| = |\Phi_v(Y)|\,|\Phi_g(Y)|$ and
$\angle Z = \angle\Phi_v(Y) + \angle\Phi_g(Y)$; a real (sigmoid) gate could only
rescale magnitude, whereas this gate can also steer phase. The output map
$W_{\mathrm{out}} = W_r + i W_i$ is a complex linear layer,
$W_{\mathrm{out}}Z = (W_r\,\mathrm{Re}\,Z - W_i\,\mathrm{Im}\,Z) + i(W_r\,\mathrm{Im}\,Z + W_i\,\mathrm{Re}\,Z)$.

\paragraph{Implicit spectral filter.}
The filter $H \in \mathbb{C}^{H\times W\times D}$ is not stored but produced by a
small MLP applied to a sinusoidal embedding of the normalised frequency grid
$(\text{row},\text{col})\in[0,1]^2$. Because $H$ is regenerated at the incoming
grid size on every forward pass, the model applies the learned filter at any
resolution without truncation, padding, or interpolation, and extrapolates to
grids unseen during training.

\paragraph{Phase-preserving primitives.}
Two operations keep the phase intact. ModReLU rectifies magnitude while leaving
the direction fixed, and the normalisation rescales by the RMS of the
magnitudes only:
\begin{equation}
\mathrm{ModReLU}(z) = \mathrm{ReLU}(|z| + b)\,\frac{z}{|z|}, \qquad
\mathrm{PhaseNorm}(z) = \frac{z}{\sqrt{\overline{|z|^2}+\epsilon}}\;\odot\,\gamma .
\end{equation}
Dropout shares one Bernoulli mask across the real and imaginary parts, so a
dropped feature loses its magnitude and phase together rather than having its
angle perturbed.

\paragraph{Readout.}
After the final phase-norm, a bridge leaves the complex domain by concatenating
the real and imaginary parts and projecting, $h = \mathrm{LN}(W_b[\mathrm{Re}\,z_L \,\|\, \mathrm{Im}\,z_L])$;
this LayerNorm is the only deliberately phase-destroying operation in the model.
A spatial mean over the grid and a linear head give the logits,
$\hat y = W_{\mathrm{head}}\,\mathrm{mean}_{H,W}\,h$.

\paragraph{Precision and variants.}
All Fourier and phase operations run in single precision while the real-valued
sublayers use bfloat16 autocast, since complex kernels have no half-precision
path. The model studied in the main text uses spatial-only mixing and ModReLU; a
channel-mixing FFT (over $H\times W\times D$) and a phase-destroying activation
(CReLU) are used only as ablations.

\begin{table}[h]
\centering
\caption{PRISM2D configuration.}
\begin{tabular}{lcccccc}
\toprule
image & patch & grid & $D$ & depth & filter MLP & params \\
\midrule
224 & 16 & $14\times14$ & 256 & 10 & 64 & $\sim$7M \\
\bottomrule
\end{tabular}
\end{table}

\section{Training Details}
\label{app:training}

PRISM2D and GFNet-Ti were trained from scratch by us on ImageNet-100
\citep{tian2020contrastive} under a single shared recipe, so that comparisons
between them isolate the complex against the real spectral design rather than a
difference in training. ResNet-50 and ViT-B/16 are intentionally off-the-shelf
ImageNet-1k \citep{imagenet15russakovsky} models (Section~\ref{sec:method}); we
include them not as capacity-matched competitors but to test whether the
phase/sign phenomenon holds across a roughly order-of-magnitude range of scale
and across architecture families we did not train.

\paragraph{Optimisation (shared).}
\begin{itemize}\itemsep2pt
  \item Optimiser: AdamW, $\beta=(0.9,0.999)$, weight decay $0.05$.
  \item Schedule: cosine decay with $5$-epoch linear warmup; peak LR $1\times10^{-3}$, min LR $1\times10^{-5}$.
  \item $100$ epochs, batch size $256$, gradient clipping at norm $1.0$.
  \item Precision: bfloat16 autocast for real-valued sublayers; all Fourier and phase operations in FP32 (complex kernels have no half-precision path).
  \item Single A100 GPU for PRISM, L4 GPU for GFNet, seed $42$.
\end{itemize}

\paragraph{Regularisation and augmentation (shared).}
\begin{itemize}\itemsep2pt
  \item RandAugment (\texttt{rand-m9-mstd0.5-inc1}), random erasing $p=0.25$, bicubic interpolation.
  \item Mixup $\alpha=0.8$ and CutMix $\alpha=1.0$; label smoothing $0.1$ (soft-target cross-entropy).
  \item ImageNet normalisation ($\mu=(0.485,0.456,0.406)$, $\sigma=(0.229,0.224,0.225)$).
  \item Evaluation: resize to $248$ (bicubic) and centre-crop to $224$.
\end{itemize}

\paragraph{Data.}
We use ImageNet-100, the $100$-class subset of ImageNet
\citep{imagenet15russakovsky} introduced by \citet{tian2020contrastive}, in the
\texttt{clane9/imagenet-100} release: $126{,}689$ training and $5{,}000$
validation images over $100$ classes at $224\times224$ resolution.

\paragraph{Architecture-specific.}
Both models use width $256$. GFNet-Ti has depth $12$ with stochastic depth
(drop-path $0.1$) and no other dropout; PRISM2D has depth $10$, complex dropout
$0.1$ (shared real/imaginary mask), and no stochastic depth. All other
hyperparameters above are identical across the two.
\end{document}